\documentclass[a4paper,fleqn]{cas-sc}

\usepackage[numbers]{natbib}
\usepackage{graphicx}
\usepackage{hyperref}
\usepackage{pgfplots}
\usepackage{xcolor}
\pgfdeclarelayer{background}
\pgfdeclarelayer{foreground}
\pgfsetlayers{background,main,foreground}
\usetikzlibrary{shapes.geometric, arrows}
\usepackage[export]{adjustbox}
\newlength\figwidth
\newlength\figheight
\tikzset{mark options={solid,mark size=3,line width=1pt,mark repeat=15}}
\pgfplotsset{every axis plot/.append style={line width=1.5pt}}
\usepackage{url}
\usepackage{array,tabularx}
\usepackage{multicol}  
\usepackage{algorithmic} 
\usepackage[boxruled,longend,linesnumbered,ruled]{algorithm2e}
\usepackage{subcaption}
\usepackage{float}
\usepackage{comment}
\DeclareGraphicsExtensions{.pdf,.png}
\def\tsc#1{\csdef{#1}{\textsc{\lowercase{#1}}\xspace}}
\tsc{WGM}
\tsc{QE}
\tsc{EP}
\tsc{PMS}
\tsc{BEC}
\tsc{DE}
\pgfplotsset{compat = 1.17}
\begin{document}
\let\WriteBookmarks\relax
\def\floatpagepagefraction{1}
\def\textpagefraction{.001}
\shorttitle{RR Estimation from PPG}
\shortauthors{Shuzan et~al.}

\title [mode = title]{A Novel Non-Invasive Estimation of Respiration Rate from Photoplethysmograph Signal Using Machine Learning Model}

\author[1]{Md Nazmul Islam Shuzan}
\ead{nazmul.shuzan@northsouth.edu}

\author[1]{Moajjem Hossain Chowdhury}
\ead{moajjem.hossain@northsouth.edu}
\author[2]{Muhammad E.H. Chowdhury}
\ead{mchowdhury@qu.edu.qa }
\author[3]{M. Monir Uddin}
\ead{monir.uddin@northsouth.edu}
\author[2]{Amith Khandakar}
\cormark[1]
\ead{amitk@qu.edu.qa}
\author[3]{Zaid B. Mahbub}
\ead{zaid.mahbub@northsouth.edu}
\author[4]{Naveed Nawaz}
\ead{nnawaz@qu.edu.qa}

\cortext[cor1]{Corresponding author}

\address[1]{Department of Electrical and Computer Engineering, North South University, Dhaka 1229, Bangladesh}

\address[2]{Department of Electrical Engineering, Qatar University, Doha 2713, Qatar}

\address[3]{Department of Mathematics and Physics, North south University, Dhaka-1229, Bangladesh}
\address[4]{Department of Computer Science and Engineering, Qatar University, Doha 2713, Qatar}


\begin{abstract}
Respiratory ailments such as asthma, chronic obstructive pulmonary disease (COPD), pneumonia, and lung cancer are life-threatening. Respiration rate (RR) is a vital indicator of the wellness of a patient. Continuous monitoring of RR can provide early indication and thereby save lives. However, a real-time continuous RR monitoring facility is only available at the intensive care unit (ICU) due to the size and cost of the equipment. Recent researches have proposed Photoplethysmogram (PPG) and/ Electrocardiogram (ECG) signals for RR estimation however, the usage of ECG is limited due to the unavailability of it in wearable devices. Due to the advent of wearable smartwatches with built-in PPG sensors, it is now being considered for continuous monitoring of RR. This paper describes a novel approach to RR estimation using machine learning (ML) models with the PPG signal features. Feature selection algorithms were used to reduce computational complexity and the chance of overfitting. The best ML model and the best feature selection algorithm combination was fine-tuned to optimize its performance using hyperparameter optimization. Gaussian Process Regression (GPR) with fitrgp feature selection algorithm outperformed all other combinations and exhibits a root mean squared error (RMSE), mean absolute error (MAE), and two-standard deviation (2SD) of 2.57, 1.91, and 5.13 breaths per minute, respectively. This ML model based RR estimation can be embedded in wearable devices for real-time continuous monitoring of the patient.
\end{abstract}




\maketitle

\section{Introduction}
\label{sec:introduction}

\hspace{\parindent}One of the most important physiological parameters that are used to diagnose abnormality in a human body is respiration rate (RR). It is one of the four primary vital signs along with heart rate, blood pressure, and body temperature. RR is expressed as the number of breaths a person takes in one minute (breaths/minute). An unusual RR is often a cause for concern and is often used as an indicator for an ailing body \cite{fieselmann1993,goldhill1999,ebell2007}. Hence, it is a vital parameter that is monitored by healthcare personnel when they check for acute deterioration of the patients \cite{cretikos2007}. Problems in the respiratory system \cite{gravelyn1980}, cardiac arrest \cite{schein1990} and even death occur- ring during hospital stay \cite{duckitt2007} can be predicted by an increased RR.  So, hospital patients who are very ill have their RR measured every few hours \cite{william2012}. Its importance is also noted in emergency departments of hospitals where they use RR for screening \cite{farrohknia2011}. Furthermore, RR is used to diagnose pneumonia \cite{karlen2013,khalil2007}and sepsis during primary treatment.\\

RR is also used to identify pulmonary embolism \cite{pimentel2015,goldhaber1999} and hypercarbia \cite{cretikos2008}. Hence, there must be an accurate way of measuring RR in clinical settings as it would greatly benefit both the patient and health care providers. However, even now it is mostly being estimated by counting the breaths manually. This method is not suitable when the patient needs to be monitored unobtrusively. It also requires more effort from the medical personnel when measuring RR. Furthermore, this method is error-prone \cite{lovett2005,philip2015} and is capnography, where the concentration of partial pressure of carbon dioxide (CO2) in the respiratory gases\cite{jaffe2008} is measured. It is one of the most accurate ways of measuring RR. However, it is cumbersome to use. As a result, it is mainly used during anesthesia and intensive care. So, alternate noninvasive methods need to be developed.\\

One of the most popular alternatives is to use either electrocardiogram (ECG) or photoplethysmogram (PPG) to estimate RR. ECG and PPG signals are easily measured during a clinical assessment. They can also be measured easily by devices for health care monitoring. Hence, there is a potential for automating the process of RR estimation without the necessity of using capnography machines. Many algorithms have been proposed for estimating RR from ECG \cite{moody1985,orphanidou2013,mir2014}. However, it has been observed that respiratory signals extracted from ECG appeared flat in ICU patients even though they were breathing sufficiently \cite{drew2014}. Besides, the clinical ECG system still requires trained professionals to operate and are bulky. Hence, the PPG signal has become more appealing for estimating RR.\\

Several recent developments on the estimation methods of the RR were comprehensively summarized in this section \cite{charltonreview2017,charlton2016,charlton2017}. A diverse range of methodologies was used to test the efficiency of RR algorithms using ECG and PPG waveform and the majority of them used PPG signals. Various issues make it difficult to reinvestigate the performance of the reported algorithms. In \cite{charlton2016,charlton2017},  about 100 algorithms have been suggested to measure the respiratory rate (RR) from ECG and PPG. All high-performance algorithms are composed of innovative variations of time domain RR estimation and modulation fusion techniques. In \cite{karlen2013}, the authors proposed a novel method for estimating the respiratory rate in real-time from the PPG signals. The incremental-merge segmentation algorithm was used to derive three respiratory-induced variations (frequency, strength, and amplitude) from the PPG signal. The smart fusion showed trends of improved estimation of root mean square error (RMSE) 3.0 breaths per min (bpm) compared to the individual estimation methods.\\

In \cite{shah2015}, the authors introduced a feasible alternative for estimating child respiratory rates during evaluation in the emergency department, particularly if the segments of PPG  contaminated by the movement artifacts were automatically discarded by an appropriate algorithm. They achieved a mean absolute error (MAE) of 5.2 bpm for the age group of 5-12 years. In \cite{zhang2017},  a novel method was proposed to estimate the RR of the PPG signal using joint sparse signal reconstruction (JSSR) and spectra fusion (SF). In \cite{motin2020}, a smart fusion method was introduced based on ensemble empirical mode decomposition (EEMD) to improve RR extraction from PPG. In \cite{motin2019}, they applied EEMD and tested on two different datasets. In\cite{l2019}, PPG-RR calculations were retrospectively conducted on PPG waveforms derived from the data warehouse and compared with RR reference values during the validation stage of the algorithm. In \cite{pirhonen2018}, the use of amplitude fluctuations of the transmittance mode finger PPG signal in RR estimation by comparing four time-frequency (TF) signal representation approaches cascaded with a particle filter was studied. \\

In \cite{jarchi2018}, a case study of 10 patients was reported for whom fewer RR estimates were derived from PPG signals relative to accelerometry. In \cite{hartmann2019}, the disparity in the precision of PPG-derived respiration frequency between measurements at various body sites for normal and deep breathing conditions was investigated. Respiratory signals were derived from PPG signals of 36 healthy subjects using the frequency demodulation method to measure respiration frequency via spectral power density. The linearity between the PPG-derived and the reference respiratory frequency was highest on the forehead. In \cite{luguern2020}, Charlton’s method \cite{charlton2016,charlton2017} was used with remote PPG (rPPG) based signals to boost the accuracy of the respiration rate estimation. Few improvements have been made to make it usable for rPPG signals. Using PPG-contact algorithms on remote PPG signals can lead to respiratory rate estimates with an MAE of less than 3 bpm and the reported MAE and RMSE Of 3.03 and 3.69 bpm, respectively.\\

 Table \textbf {Table \ref{Literature Review}},summarizes a wide variety of RR estimation algorithms from the PPG that have been published in recent years. None of them used machine learning (ML) models to estimate the RR from ECG or PPG and their fusion. Therefore. There is a potential scope to use ML models to improve the RR estimation algorithm. With the increase of the availability of annotated datasets, it is possible to use ML techniques in RR algorithms \cite{charltonreview2017}, which is a major motivation of this study. However, to the best of our knowledge, no recent work has derived t- domain, f-domain, and statistical features from PPG signal to estimate reliably RR using the machine learning models. In our previous studies \cite{chowdhury2020,chowdhury2019,chowdhury2019real}, several time-domain features were calculated from the original signal and its derivatives. Several features were extracted for RR estimation from the PPG signal in this study, which was not used before by any other research group.\\
 
 This manuscript is divided into four sections where \textbf{Section \ref{sec:introduction}} addresses the fundamentals of the PPG signal, the associated works, and the motivation for this study. The database description, pre-processing, evaluation measures, and methodology are discussed in \textbf{Section \ref{sec:Methodology}} while \textbf{Section \ref{Results and Discussion}} outlines the results and discusses them, and compares them with some other research solutions, while, while \textbf{Section \ref{Conclusions}} concludes the work.
\begin{table}[H]
\centering
\caption{Summary of methods for respiration rate estimation}
\begin{adjustbox}{max width=1\textwidth,center}
\begin{tabular}{ c| c}
 \textbf{Author} & \textbf{Method Used}  \\ 
\hline
Karlen et al.\cite{karlen2013} & Fast Fourier Transformation (FFT)  \\  
\hline
Orphanidou et al.\cite{orphanidou2013} & Ensemble Empirical Mode Decomposition \\  
\hline
Pimentel et al.\cite{pimentel2015} & Auto-regressive Model \\  
\hline
Philip et al.\cite{philip2015} & Spot Assessment \\  
\hline
Mirmohamadsadegh et al.\cite{mir2014} &  Instantaneous Frequency Tracking Algorithm \\  
\hline
Lin et al.\cite{lin2017} &  Wavelet-Based Algorithm \\  
\hline
Fleming et al.\cite{fleming2007} &  Auto-regressive Model \\
\hline
Zhou et al.\cite{zhou2006} & Independent Component Analysis (ICA) Algorithm \\
\hline
Moreno et al.\cite{moreno2018} & Digital Filtering \\
\hline
Nilsson et al.\cite{nilsson2000} & Digital Filtering \\
\hline
Motin et al.\cite{motin2019,motin2020} & Empirical Mode Decomposition  \\  
\hline
Jarchi et al.\cite{jarchi2018} & Accelerometer Based  \\  
\hline
Hartmann et al.\cite{hartmann2019} & Fast Fourier Transformation (FFT)  \\  
\hline
Pirhonen et al.\cite{pirhonen2018} & Wavelet-Based  \\  
\hline
Zhang et al.\cite{zhang2017} & Joint Sparse Signal Reconstruction\\  
\hline
\end{tabular}
\label{Literature Review}
\end{adjustbox}
\end{table}
\section{Methodology}
\label{sec:Methodology}
\hspace{\parindent}This section summarizes the dataset description and the preprocessing techniques, various features that were extracted, different feature selection algorithms, and the different machine learning models that were implemented for RR estimation in this study.\\

\begin{figure}[pos = H]
\centering
\begin{adjustbox}{max width=1\textwidth,center}   
\includegraphics{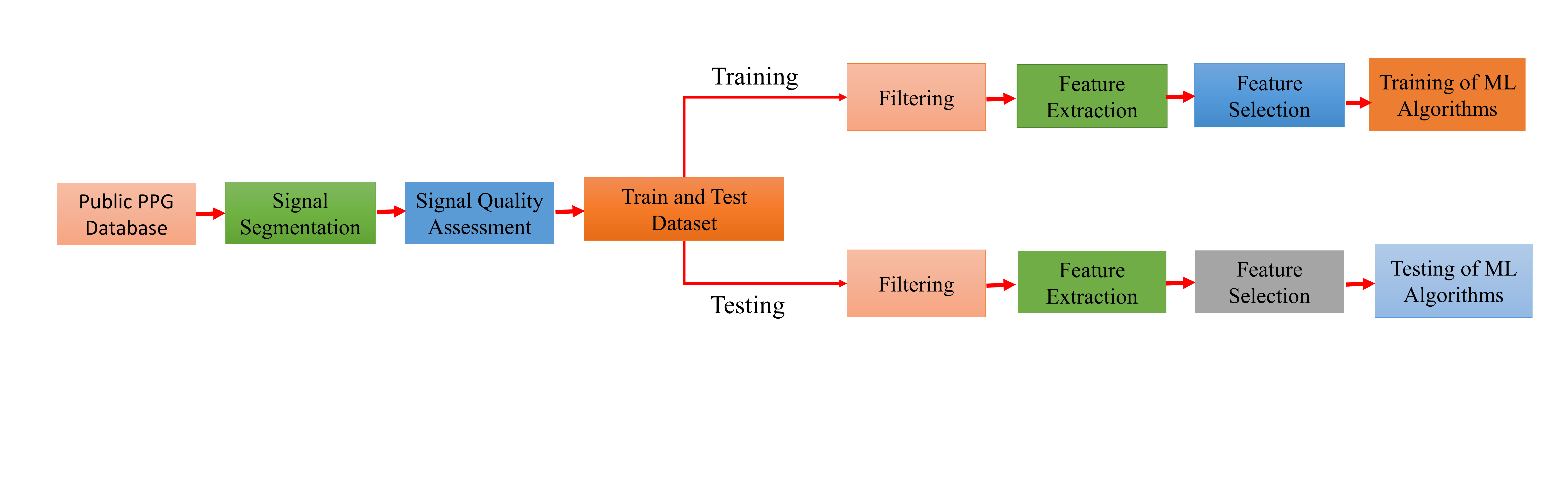}
\end{adjustbox}
\caption{Overview of the process}
\label{Overall Flowchart}
\end{figure}

\textbf{Figure \ref{Overall Flowchart}} shows the overall methodology where PPG signal from the publicly available VORTAL dataset \cite{charlton2016,charlton2017} is first segmented into windows of 32 seconds. The quality of the segmented signals were evaluated and unfit data were rejected( details in the section below). The acceptable signals are then split into 80\% training and 20\% test sets, respectively for 5-fold cross-validation. Firstly, the segmented PPG signals were filtered and meaningful features were extracted and feature selection algorithms were used to reduce feature dimensions to avoid the risk of overfitting and to reduce the computation time. The selected features were used to train, validate and test machine learning models. An unseen 20\% test-set was used to predict the RR value from the PPG features.

\subsection{Dataset Description}

\hspace{\parindent}Electrocardiogram (ECG) and photoplethysmogram (PPG) signals and the respiration rate (RR) from 39 subjects are available in the VORTAL dataset. The PPG signals used were acquired during the resting period and sampled at 500Hz sampling frequency. The summary of the dataset is shown in \textbf{Table \ref{Data Descriptor}}.

\begin{table}[H]
\centering
\caption{Characteristics of the subjects in Vortal Dataset}
\begin{tabular}{ c| c c c}
& median & Lower quartile & Upper quartile  \\ 
\hline
Sex (female) & 54\% & - & - \\  
Age (years) & 29 & 26 & 32 \\
BMI (($kg/m^2$)) & 23 & 21 & 26 \\
Respiration Rate (bpm) &5-32 & - & - \\
\hline
\end{tabular}
\label{Data Descriptor}
\end{table}

The signals were segmented into windows of 32 seconds as it allows a sufficient amount of breaths to take place so that RR can be calculated reliably \cite{charlton2016,charlton2017,karlen2013}. A shorter window will pose a problem to the respiration rate while the longer window will not be practically feasible. 761 PPG segments of 32-seconds were obtained. Thirty-one segments were rejected and 730 signals were accepted after quality evaluation. The rejected signals did not have identifiable fiducial points, had spikes, and many other deformities. Figure 2 shows the sample accepted and rejected PPG signals.

\begin{figure}[pos = H]
    \begin{center}
   \begin{adjustbox}{max width=01\textwidth,center}
  \includegraphics{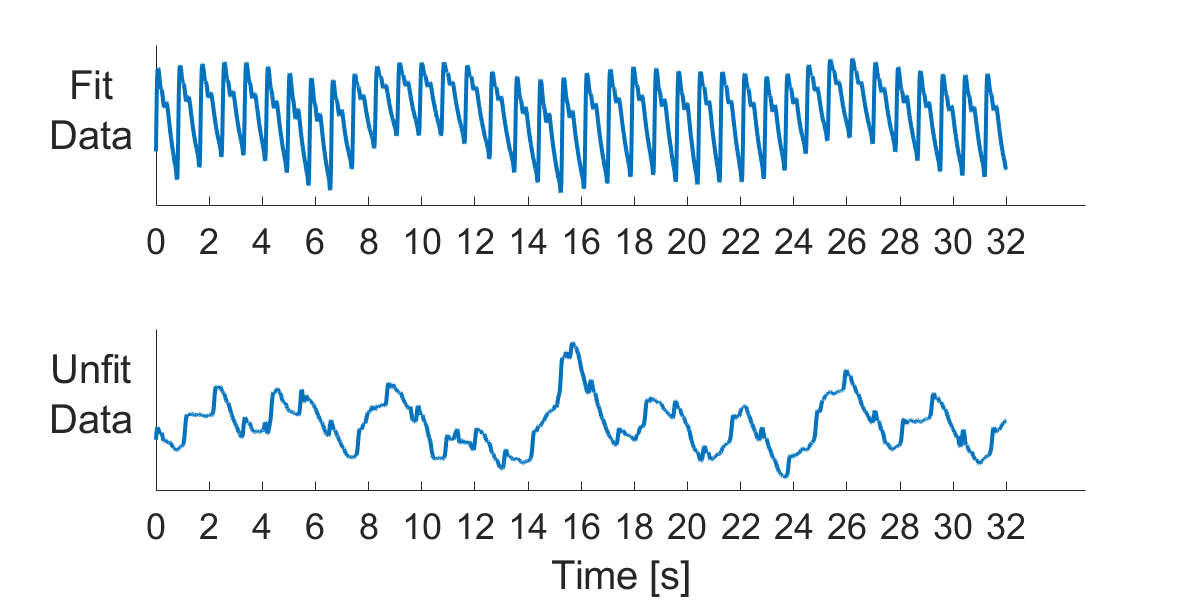}
   \end{adjustbox}
   \end{center}
    \caption{Comparison between Fit and Unfit PPG Waveforms.}
    \label{fit and unfit}
\end{figure}

\subsection{Preprocessing}
\begin{figure}[pos = H]
\centering
\begin{adjustbox}{max width=0.8\textwidth,center}
\input{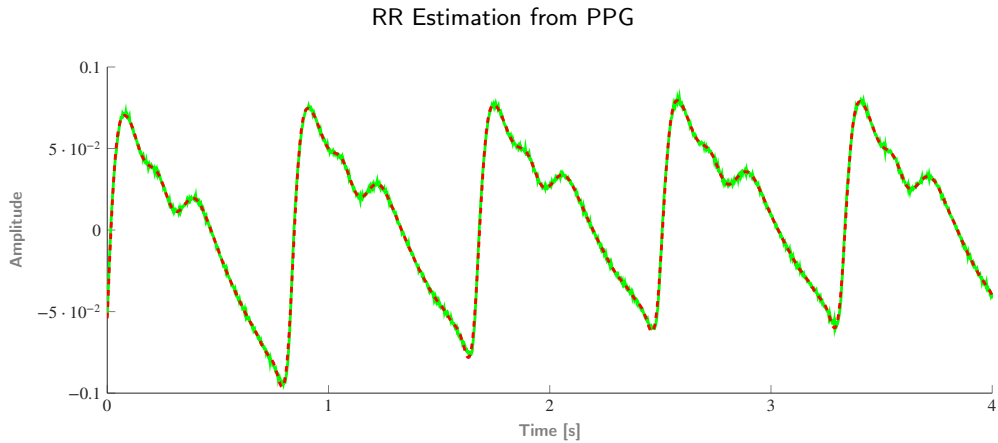}
\end{adjustbox}
\caption{Filtered Signal overlaid on the raw PPG signal.}
\label{filter}
\end{figure}

\hspace{\parindent}The PPG waveform in the dataset has high-frequency noise components. These noises can hamper the feature extraction process. Therefore, the PPG waveforms were filtered through a low-pass Butterworth Infinite Impulse Response (IIR) Zero-Phase Filter \cite{chatterjee2018}. \textbf{Figure \ref{filter}} shows the raw PPG signal overlaid with the filtered signal. A sixth-order IIR filter with a cut-off frequency of 25 Hz was implemented in MATLAB.
In the pre-processing steps, the baseline wandering present in the PPG waveform was not removed. This is because the baseline wandering present in the PPG signal is due to the effect of breathing, which we would like to extract.

\subsection{Feature Extraction}
\begin{figure}[pos = H]
\begin{center}
\begin{adjustbox}{max width=0.75\textwidth,center}
\includegraphics{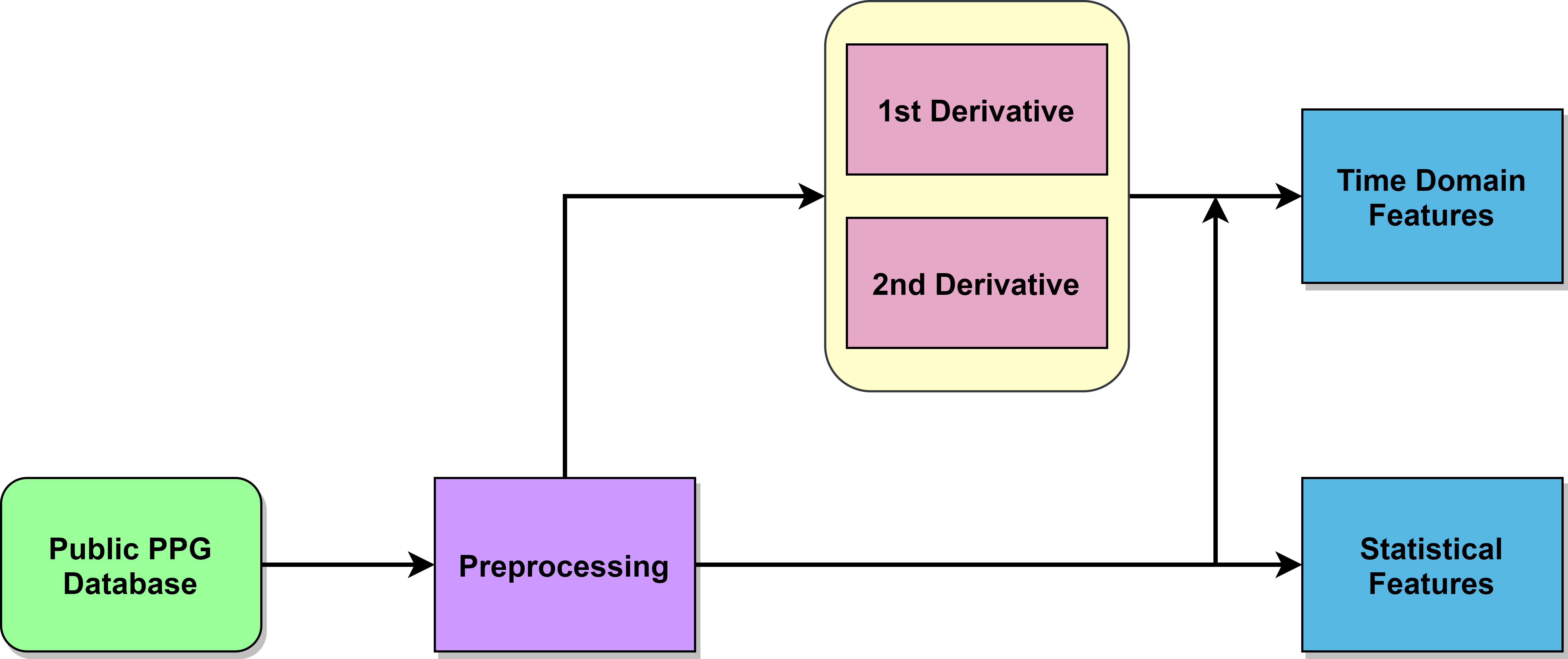}
\end{adjustbox}
\end{center}
\caption{Overview of the Feature Extraction method.}
\label{feat flowchart}
\end{figure}

\hspace{\parindent}\textbf{Figure \ref{feat flowchart}} summarizes different types of features extracted in this study. PPG waveforms are rich in detail and contain many features of interest. They contain features such as systolic peak, foot of the waveform, pulse width, peak-to-peak interval, etc. To extract the meaningful features, as shown in \textbf{Table \ref{Feature Extraction 1}}, we used the feature extraction techniques described in \cite{chowdhury2020}.\\

The preprocessed signal is used to calculate statistical features while the time-domain features were extracted from the PPG signal and its 1st and 2nd derivatives (In \textbf{Figure \ref{ppg derivative}}) From the derivatives of the signal, the main features were the first peak and first trough of the signal. Time and amplitude features were calculated afterward and summarized in \textbf{Table \ref{Feature Extraction 1}} and \textbf{\ref{Feature Extraction 2}}. Mean, standard deviation, and variance of most of the time-domain features was also calculated. This is because to capture the distortion and modulation caused by breathing on PPG, these features are important. These time-domain features were identified from different previous works \cite{chowdhury2020,chowdhury2019,chowdhury2019real}. Statistical features used in this work were identified from \cite{chowdhury2019}. In total, 107 features were extracted to feed the machine learning models.

\begin{figure}[pos = H]
    \centering
    \begin{adjustbox}{max width=1\textwidth,center}
    \includegraphics{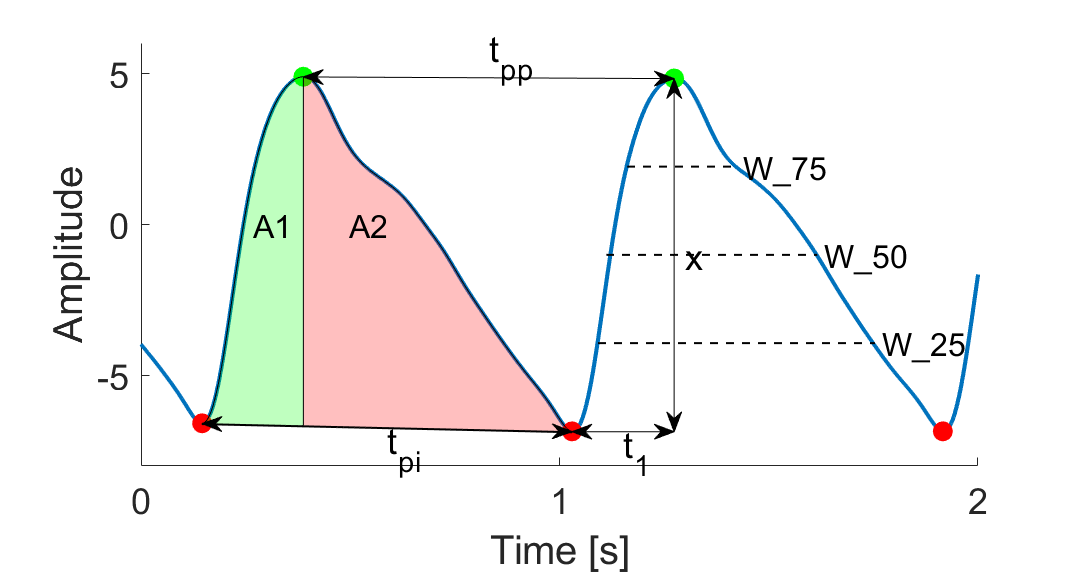}
   \end{adjustbox}
    \caption{PPG signal with some time-domain features.}
    \label{ppg feat}
\end{figure}

\begin{figure}[pos = H]
    \centering
    \begin{adjustbox}{max width=0.85\textwidth,center}
    \input{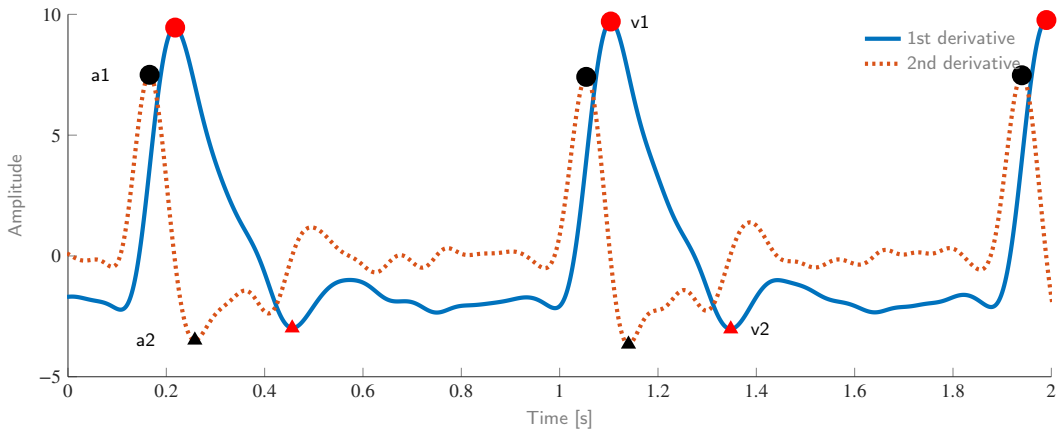}
    \end{adjustbox}
    \caption{1st and 2nd derivatives of PPG Signal.}
    \label{ppg derivative}
\end{figure}

\begin{table}[pos = H]
\centering
\caption{Fifteen time-domain features with their mean, standard deviation and variance}
\begin{adjustbox}{max width=1\textwidth,center}
\begin{tabular}{ c| c }
 \textbf{Features} &\textbf{Definition} \\ 
\hline
Systolic Peak & The amplitude of (‘sys’) from the PPG signal \\ 
\hline
Height of foot & The amplitude of (‘amp foot’) from the PPG signal \\ 
\hline
New Systolic peak & The amplitude of (‘x’) from the PPG signal\\ 
\hline
Systolic peak time & The time interval from the foot of the PPG signal to the systolic peak (‘t1’) pulse\\
\hline
Interval & The time interval from foot to next foot (‘tpi’)\\
\hline
Peak-to-Peak Interval & The time distance between two consecutive systolic peaks (‘tpp’) \\
\hline
t1/x & The ratio of systolic peak time to the systolic amplitude of the PPG waveform\\
\hline
t1/tpi & The ratio of systolic peak time to pulse interval of the PPG waveform \\
\hline
x/(tpi-t1) & The ratio of ’x’ to the difference between ‘tpi’and ‘t1’\\
\hline
Rising area & Area from first foot to systolic peak(‘A1’)\\
\hline
Decay area & Area from systolic peak to foot(‘A2’) \\
\hline
A1/A2 & The ratio from ‘A1’ to ‘A2’\\
\hline
Width (25\%) & The width of the PPG signal at 25\% amplitude of ‘x’ \\
\hline
Width (50\%) & The width of the PPG signal at 50\% amplitude of ‘x’\\
\hline
Width (75\%) & The width of the PPG signal at 75\% amplitude of ‘x’ \\
\hline

\end{tabular}
\end{adjustbox}
\label{Feature Extraction 1}
\end{table}

\begin{table}[pos = H]
\centering
\caption{Sixteen features derived from the first and second derivative with mean, standard deviation, and variance}
\begin{adjustbox}{max width=1\textwidth,center}
\begin{tabular}{ c| c }
\textbf{Features} &\textbf{Definition} \\ 
\hline
v1 & The first maximum peak from the 1st derivative of the PPG signal  \\ 
\hline
tv1 & The first maximum peak time from the 1st derivative of the PPG signal  \\ 
\hline
v2 & The first minimum peak amplitude from the first derivative \\ 
\hline
tv2 & The first minimum peak time from the first derivative of the PPG signal \\
\hline
a1 & The first maximum peak amplitude from the 2nd derivative of the PPG signal \\
\hline
ta1 & The first maximum peak time from the 2nd derivative of the PPG signal \\
\hline
a2 & The first minimum peak amplitude from the second derivative of the PPG signal \\
\hline
ta2 & the first minimum time amplitude from the second derivative of the PPG signal\\
\hline
v2/v1 & The ratio between v2 and v1 \\
\hline
a2/a1 & The ratio between a2 and a1 \\
\hline
tv1/tv2 & The ratio between tv1 and tv2\\
\hline
ta1/ta2 & The ratio between ta1 and ta2  \\
\hline
tv1/ta1 & The ratio between tv1 and ta1  \\
\hline
tv1/ta2 & The ratio between tv1 and ta2  \\
\hline
tv2/ta1 & The ratio between tv2 and ta1  \\
\hline
tv1/ta1 & The ratio between tv1 and ta2 \\
\hline
\end{tabular}
\end{adjustbox}
\label{Feature Extraction 2}
\end{table}
\subsection{Feature Selection}
\hspace{\parindent}Feature selection decreases the data dimensionality by choosing only a subset of calculated characteristics (predictor variables) to construct a model. Feature selection algorithms (FSA) look for a subset of predictors that optimally model the responses tested, considering the constraints such as feature importance and subset size. The Feature Ranking Library (FSLib) is an often-used MATLAB library \cite{roffo2016}. In this work, 10 Feature selection algorithms have been used and after several feature combinations, the best feature ranking technique for this problem is identified.\\

\underline{Fit a Gaussian process regression model (Fitrgp)}: Fitrgp can find the predictor weights by taking the exponential of the negative learned length scales contained in the kernel information property \cite{rasmussen2003,lagarias1998}. In \textbf{Table \ref{fsa1}}, it is found that the most contributory features are 9, out of 107 features and 6 among the 9 selected features are derived from the derivative of the PPG signal.\\
 
\underline{Least absolute shrinkage and selection operator (Lasso)}: Lasso minimizes the variance of inference by retaining the sum of the absolute values of the model parameters smaller than the fixed value \cite{liu2007}. The most contributory features are 25, out of 107 features where 13 features were derived from the PPG signal (\textbf{Table \ref{fsa1}}).\\
 
\underline{Relieff feature selection (RFS)}: RFS works much better to approximate the significance of the function for distance-based supervised models that use pairwise distances between the observations to predict \cite{kononenko1997,robnik2003}. \textbf{Table \ref{fsa1}}, shows that 10 features are the most contributory features and 6 out of them are PPG signal features.\\
 
\underline{Feature selection with adaptive structure learning (Fsasl)}: Fsasl is focused on linear regression and its only limitation is the high computational complexity, which can be expensive for the high-dimensional results \cite{du2015}. \textbf{Table \ref{fsa2fs}}, shows the most contributory 15 features where 9 are the PPG signal features.\\
 
\underline{Unsupervised feature selection with ordinal locality (Ufsol)}: To implement the selected feature classes, a triplet-based loss function is added to maintain the ordinal localization of original data, which leads to distance-based clustering activities. And then simplify orthogonal base clustering by imposing an orthogonal restriction on the function projection matrix. As a consequence, a general structure for simultaneous collection and clustering of features is addressed \cite{guo2017}. \textbf{Table \ref{fsa2fs}}, shows the most contributory 17 features, where 8 features are from the PPG signal and its derivative.\\

\underline{Laplacian method (LM)}: Another unsupervised approach is the LM, where the value of a feature is determined by its capacity to conserve the locality. This approach builds the closest neighbor graph to model the local geometric structure. LS algorithm is searching for features that respect the structure of this graph \cite{he2006}. \textbf{Table \ref{fsa2fs}}, lists the most contributory 10 features where  8 features were extracted from the PPG signal.\\
 
\underline{Unsupervised dependence feature Selection (UDFS)}: UDFS is a projection-free function selection model based on l2.0-standard equality constraints. UDFS conducts the collection of function subsets by optimizing two terms: one term increases the dependency on the original results, while the other term maximizes the dependence of selected features on cluster labels to direct the phase of subset feature selection \cite{yang2011}. It was found that 8 out of 10 most contributory features were from the PPG signal (\textbf{Table \ref{fsa2fs}}).\\
 
\underline{Infinite Latent feature selection (ILFS)}: ILFS is a probabilistic approach to latent feature selection that performs the ranking stage by taking into consideration all feasible sub-sets of features that circumvent the combinatorial issue \cite{zeng2010feature}. \textbf{Table \ref{fsa3}}, shows the top-ranked 20 features among which 10 were from the PPG signal.\\
 
\underline{Multi cluster feature selection (mCFS)}: mCFS requires a sparse eigenproblem and an L1- regularized least squares question to efficiently solve the corresponding optimization problem \cite{cristani2012con}. Top-ranked 15 features were identified where 8 features were contributed by the derivative of the PPG signal.\\
 
\underline{Correlation based feature selection (CFS)}: CFS is an embedded process that selects features in a sequential backward exclusion fashion to rank top features using linear SVM \cite{guyon2002gene}. In \textbf{Table \ref{fsa3}}, the most contributory features are 15 of the 107 features and 9 of the 15 features contributed only on the PPG signal derivatives.

\begin{table}[pos = H]
\centering
\caption{Top-ranked features using Fitrgp, Lasso, and Relieff based Feature Selection Algorithms}
\begin{adjustbox}{max width=1\textwidth,center}
\begin{tabular}{||m{0.15\textwidth}| m{0.22\textwidth} m{0.22\textwidth} m{0.22\textwidth} ||}
\hline
\textbf{Algorithm}&\multicolumn{3}{|c|}{\textbf{Best Selected Features}}\\ [0.5ex] 
 \hline\hline
 \multirow{3}{*}{Fitrgp (9)} 
 &mean(tv1) & mean(tv2/ta1)&mean(v2/v1) \\
 &mean(tpp)&mean(ta1/ta2)&std(tpi)\\
 &mean(x/(tpi-t1))&mean(v2/v1)&mean(ta1)\\
 \hline
 \multirow{9}{*}{Lasso (25)}
&mean(v1)&mean(sys)&mean(tpp)\\
&iqr(sig)&mean(A2)&std(A2)\\
&mad(sig)&mean(foot amp) &mean(t1)\\
&mean(x)&mean(A1)&mean(tv2/ta1)\\
&var(w\_25)&mean(a1)&mean(tpi)\\
&mean(w\_25)&mean(a2)&mean(ta1)\\
&std(sig)&mean(ta2)&mean(tv1/ta2)\\
&25\% quantile &mean(tv1)&\\
&75\% quantile&std(A1)&\\
\hline
\multirow{4}{*}{Relieff (10)}
&mean(t1)&mean(tpi)&mean(tv1)\\
&mean(v2/v1)&mean(t1/tpi)&mean(w\_25)\\
&mean(ta2)&mean(ta1/ta2)&\\
&mean(tpp)&std(t1)&\\
\hline
\end{tabular}
\end{adjustbox}
\label{fsa1}
\end{table}
\begin{table}[pos = H]
\centering
\caption{Top-ranked features using Fsasl, Ufsol, Laplacian and UDFS based Feature Selection Algorithms}
\begin{adjustbox}{max width=1\textwidth,center}
\begin{tabular}{||m{0.17\textwidth}| m{0.22\textwidth} m{0.22\textwidth} m{0.20\textwidth} ||}
\hline
\textbf{Algorithm}&\multicolumn{3}{|c|}{\textbf{Best Selected Features}}\\ [0.5ex] 
 \hline\hline
 \multirow{5}{*}{Fsasl(15)} 
 &mean(a2) & max ratio&mean(tv2) \\
 &var(t1/x)&mean(t1/x))&std(t1/x)\\
 &mean(x)&mean(tpi)&mean(ta1/ta2)\\
 &mean(v2)&mean(tv1/tv2)&std(w\_50)\\
 &std(a2)&var(A2)&std(x)\\
 \hline
 \multirow{6}{*}{Ufsol (17)} 
 &75\%quantile& mean(a2)&mean(t1) \\
 &mean(x)&mean(ta2)&mean(tv2/ta1)\\
 &mean(A2)&mean(tv1)&mean(tpi)\\
 &mean(foot amp)&std(A1)&mean(ta1)\\
 &mean(A1)&mean(tpp)&mean(tv1/ta2)\\
 &mean(a1)& std(A2)&\\
 \hline
 \multirow{4}{*}{Laplacian(10)} 
 &std(sig)& mean(A2)&mean(x) \\
 &mad(sig)&iqr(sig)&std(A2)\\
 &mean(x)&mean(v1)&\\
 &mean(A1)&mean(foot amp)&\\
  \hline
 \multirow{5}{*}{UDFS (15)} 
 &iqr(sig)&entropy(sig)&mean(A2) \\
 &mean(w\_50)&mad(sig)&std(A2)\\
 &var(w\_50)&std(sig)&mean(A1)\\
 &25\% quantile&mean(foot amp)&mean(t1)\\
 &75\% quantile&mean(x/(tpi-t1))&mean(v1)\\
 \hline 
 \end{tabular}
 \end{adjustbox}
\label{fsa2fs}
\end{table}
\begin{table}[pos = H]
\centering
\caption{Top-ranked features using IlFS, mCFS and CFS based Feature Selection Algorithms}
\begin{adjustbox}{max width=1\textwidth,center}
 \begin{tabular}{||m{0.15\textwidth}| m{0.22\textwidth} m{0.22\textwidth} m{0.22\textwidth} ||}
\hline
\textbf{Algorithm}&\multicolumn{3}{|c|}{\textbf{Best Selected Features}}\\ [0.5ex] 
 \hline\hline
 \multirow{7}{*}{IlFS (20)} 
 &var(tv2) &mean(ta2)&mean(tv1) \\
 &maxfreq&mean(v2/v1)&std(w\_25)\\
 & mean(w\_75)&maxratio&mean(w\_25)\\
 &var(w\_75)&spectral-entropy&var(w\_25)\\
 &mean(a2/a1)& std(t1/tpi)&std(w\_50)\\
 &std(w\_75)&mean(ta1)&mean(w\_50)\\
 &entropy(sig)&mean(tv1)&var(w\_50)\\
 \hline 
 \multirow{5}{*}{mCFS (15)} 
 &mean(x) &mean(foot amp) &mean(tv2) \\
 &mean(v1) & mad(sig) &mean(tv1)\\
 &mean(A2) &mean(a1) &mean(ta1)\\
 &mean(A1)&std(sig)&mean(tv1/ta2)\\
 &mean(sys))&mean(a2) &mean(tv2/ta2)\\
 \hline 
 \multirow{5}{*}{CFS (15)} 
 &var(sys) &var(a2) &var(tv2) \\
 &mean(a2/a1) & var(ta1/ta2) &std(tpp)\\
 &var(v2) &std(sig) &mean(ta1)\\
 &median(sig)&mean(tpi)&entropy(sig)\\
 &mean(tv1/tv2)&var(a2/a1) &std(tv1)\\
 \hline
 \end{tabular}
 \end{adjustbox}
\label{fsa3}
\end{table}
\subsection{Machine Learning}
\hspace{\parindent}Training, validation, and testing of the machine learning models were performed using 5-fold cross-validation. \textbf{Table \ref{Data Set}} summarizes the number of PPG signal segments were used for training, validation, and testing. 80\% of 730 recordings were used for training while 20\% out of training samples were used for validation and 20\% of 730 recordings were used for testing.  We then extracted the features.  Regression Learner App of MATLAB 2019b \cite{regapp2019b} was used to estimate respiration rate (RR). Five different algorithms (Support Vector Regression (SVR), Gaussian Process Regression (GPR), Ensemble Trees Linear Regression, and Regression Trees) with 19 different variations were evaluated. Furthermore, Artificial Neural Network (ANN), and Generalized Regression Neural Network (GRNN) were also investigated.
\begin{table}[pos = H]
\centering
\caption{Description of train, validation, and test set}
\begin{tabular}{| c| c| c |}
\hline
Train set & Validation set & Test set \\ 
\hline
468 & 116 & 146\ \\ 
\hline
\end{tabular}
\label{Data Set}
\end{table}

\underline{Gaussian Process Regression (GPR)}: GPR is a Bayesian regression approach, which works well on small datasets. Where most of the supervised machine learning algorithms learn the exact values of the function for each parameter, GPR learns a distribution of probability over all possible values \cite{gpr}.\\

\underline{Ensemble Trees}: In this algorithm, multiple regression trees are combined using a weighted combination. The main idea behind this type of model is to use the strength of multiple weak learners to create a strong learner \cite{ensemble}.\\

\underline{Support Vector Regression (SVR)}: It is a supervised learning algorithm where SVR is trained using the symmetrical loss function that punishes both higher and lower misprediction \cite{svr}.\\

\underline{Artificial Neural Network (ANN)}: ANN tries to understand the relations between a set of data in a way that mimics the process of human brain behavior. It uses a set of interconnected artificial neurons in the layered structure and ANN can work very well on different types of data using this layered structure \cite{ann}.\\

\underline{Generalized Regression Neural Network (GRNN)}: GRNN is a special type of neural network architecture where it has a radial basis layer and a special linear layer \cite{grnn}. The uniqueness of GRNN is that it does not require a repeated training procedure like back-propagation networks compared to ANN where back-propagation is vital.

\subsection{Hyperparameter Optimization}
\hspace{\parindent}WInitial training of machine learning algorithms was carried out using the default parameters of Regression Learner App of MATLAB 2019b \cite{regapp2019b}. The performance of these machine learning algorithms can be increased by tuning or optimization of the hyper-parameters of the algorithm. Bayesian Optimization was used in this work which was tuned for 60 iterations.
\subsection{Evaluation Criteria}
\hspace{\parindent}To assess the performance of the machine learning models in this study, five performance matrices were used. Here, $X_p$ is the data that was predicted while X is the ground truth data and n denotes the number of samples or recordings.
\renewcommand{\theenumi}{\Roman{enumi}}
\begin{enumerate}
\item Mean Absolute Error (MAE): The Mean Absolute Error is the mean of the absolute of the predicted errors.
\begin{align}
    MAE = \frac{1}{n} \sum_n \lvert X_p - X \rvert
\end{align}   

\item Root Mean Squared Error (RMSE): RMSE measures the standard deviation of the prediction error or residuals, where residuals measure the distance of data points from the regression line. Therefore, RMSE is a way of measuring the spread of residuals, and the smaller the spread, the better the model.
\begin{equation}
    RMSE = \sqrt{\frac{\sum {\lvert X_p - X \rvert}^2}{n}}
\end{equation}

\item Correlation Co-efficient (R): R is used to measure how closely two variables (prediction and ground truth) are related. It is a statistical technique that also tells us how close the prediction matches with the ground truth.
\begin{equation}
    R = \sqrt{1 - \frac{MSE(Model)}{MSE(Baseline)}}
\end{equation}
\begin{center}
   $ where, MSE(Baseline) = \frac{\sum \lvert X - mean(X) \rvert}{n}$    
\end{center}

\item 2SD : Standard deviation(SD) is a statistical technique that measures the spread of data relative to its mean.  It is calculated by computing the square root of the variance. 2SD is the double of SD. 2SD is important because it represents the 95\% confidence interval.
\begin{equation}
    2SD = 2\times SD = 2\times \sqrt{\frac{\sum {(error - mean(error))}^2}{n}}
\end{equation}
\begin{center}
   $ where, error = X_p - X$    
\end{center}
\item Limit of Agreement(LOA):Limit of agreement calculates the interval in which a percentage of the differences between two measurements (prediction and ground truth) lie.  LOA captures both random (precision) and systematic (bias).  It is therefore a useful way of measuring the performance of ML models. 95\% LOA were computed in this study.

\end{enumerate} 
Among these criteria, RMSE and 2SD were chosen as the main criteria based on the reporting in the literature \cite{charltonreview2017,charlton2016,charlton2017,zhang2017}.

\section{Results and Discussion}
\label{Results and Discussion}
\begin{table}[pos = H]
\centering
\caption{Comparative performance of different machine learning models with different feature selection techniques}
\begin{adjustbox}{max width=1\textwidth,center}
 \begin{tabular}{||m{0.18\textwidth} |m{0.15\textwidth}| m{0.15\textwidth}| m{0.15\textwidth}| m{0.13\textwidth}||}
\hline
\textbf{Algorithm}& \textbf{Metric}& \textbf{GPR}& \textbf{SVR}& \textbf{Ensemble}\\ [0.5ex] 
\hline\hline
\multirow{2}{*}{All Features}
 &RMSE &2.94&3.15&3.39\\
  &2SD &5.89&6.33 &6.68\\
  \hline
  \multirow{2}{*}{Relieff}
 &RMSE &2.81&3.12&3.56\\
 &2SD &5.62&6.24 &7.02\\
 \hline
 \multirow{2}{*}{Laplacian}
 &RMSE &4.51&4.72&4.94\\
 &2SD &9.08&9.45 &9.79\\
 \hline
 \multirow{2}{*}{mCFS}
 &RMSE &3.08&3.37&3.93\\
 &2SD &6.13&6.69 &7.73\\
 \hline
 \multirow{2}{*}{UDFS}
 &RMSE &3.69&4.04&4.41\\
 &2SD &7.38&8.07 &8.71\\
 \hline
 \multirow{2}{*}{Llcfs}
 &RMSE &2.93&3.17&3.68\\
 &2SD &5.86&6.34 &7.24\\
 \hline
 \multirow{2}{*}{CFS}
 &RMSE &3.20&3.58&4.02\\
 &2SD &6.40&7.14 &7.91\\
 \hline
 \multirow{2}{*}{Fsasl}
 &RMSE &3.00&3.32&3.79\\
 &2SD &6.01&6.63 &7.46\\
 \hline
 \multirow{2}{*}{Ufsol}
 &RMSE &2.78&2.97&3.68\\
 &2SD &5.57&5.94 &7.24\\
 \hline
 \multirow{2}{*}{Lasso}
 &RMSE &2.81&2.96&3.52\\
 &2SD &5.61&5.91 &6.90\\
 \hline
  \multirow{2}{*}{Fitrgp}
 &RMSE &\textbf{2.61}&\textbf{2.90}&\textbf{3.53}\\
 &2SD &\textbf{5.19}&\textbf{5.79} &\textbf{6.92}\\
 \hline
 \end{tabular}
 \end{adjustbox}
\label{result fsa}
\end{table}

\hspace{\parindent}This section describes the evaluation results of the different machine learning algorithms used in this work. Out of the 19 classical machine learning algorithms evaluated in this study, SVR, GPR, and Ensemble trees were outperformers.\\

In \textbf{Table \ref{result fsa}}, it can be seen that the features selected by the Fitrgp technique were outperforming for different algorithms (SVR, GPR, and Ensemble Trees). This feature selection algorithm produced the best results for each ML model. However, the GPR model in combination with the Fitrgp feature selection technique provides superior performance with the state-of-the-art RMSE and 2SD of 2.61 and 5.19, respectively.\\

\begin{figure}[pos = H]
    \centering
    \begin{adjustbox}{max width=0.75\textwidth,center}
    \includegraphics{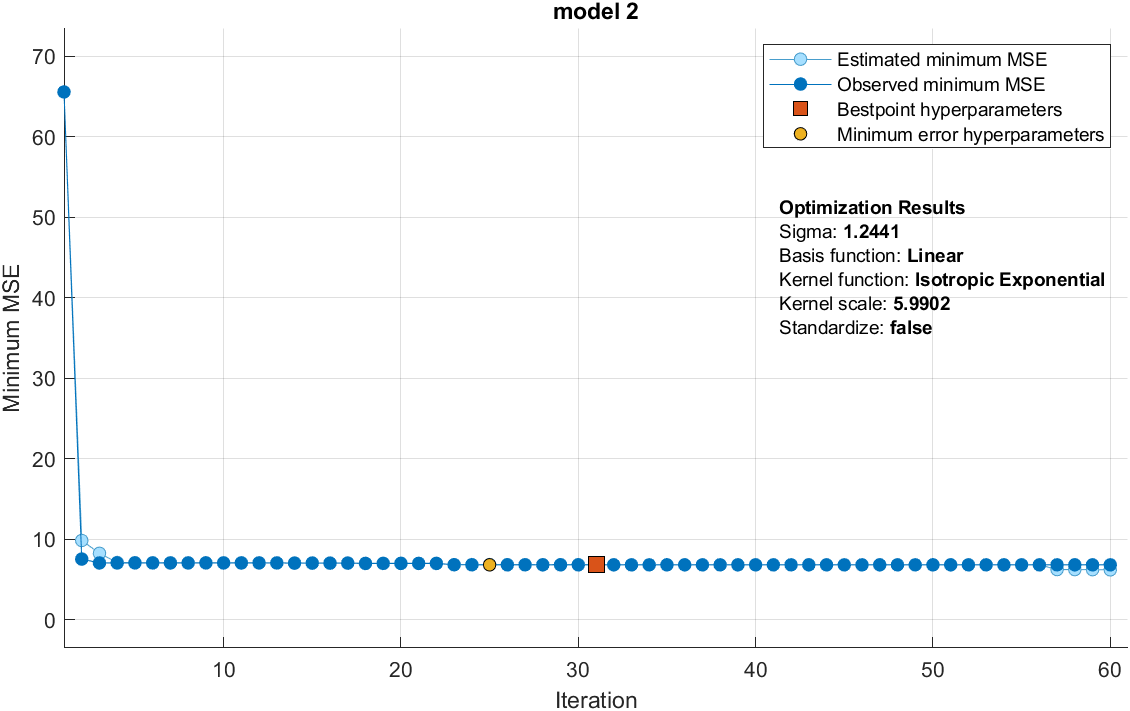}     
    \end{adjustbox}
    \caption{Hyperparameter Optimization of GPR model.}
    \label{hyper parameter}
\end{figure}

Since it has been observed in \textbf{Table \ref{result fsa}} that GPR performed the best among all classical machine learning techniques tested in this work, the hyper-parameter optimization performance of GPR was compared with ANN and GRNN. The process can be seen in  \textbf{Figure \ref{hyper parameter}}. The best model having a Sigma of 1.2441, a linear basis function, an isotropic exponential kernel function and a kernel scale of 5.9902.\\

The comparative performance of ANN, GRNN, and optimized GPR is shown in \textbf{Table \ref{result fsa2}}. It can be seen that the Fitrgp is outperforming the rest of the feature selection techniques. Among the machine learning algorithms, the optimized GPR marginally outperforms GRNN while performing significantly better than ANN. Therefore, the optimized GPR model was selected as the best performing model in this work.\\
\begin{table}[pos = H]
\centering
\caption{Performance comparison of optimized machine learning models using different feature selection techniques}
\begin{adjustbox}{max width=1\textwidth,center}
 \begin{tabular}{||m{0.18\textwidth} |m{0.15\textwidth}| m{0.15\textwidth}| m{0.15\textwidth}| m{0.13\textwidth}||}
\hline
\textbf{Algorithm}& \textbf{Metric}& \textbf{Optimized GPR}& \textbf{ANN}& \textbf{GRNN}\\ [0.5ex] 
\hline\hline
\multirow{2}{*}{All Features}
 &RMSE &2.84&4.00&3.29\\
  &2SD &5.70&7.96 &6.59\\
  \hline
  \multirow{2}{*}{Relieff}
 &RMSE &2.84&3.79&3.04\\
 &2SD &5.69&7.56 &6.09\\
 \hline
 \multirow{2}{*}{Laplacian}
 &RMSE &4.25&5.07&5.07\\
 &2SD &8.42&10.13 &10.14\\
 \hline
 \multirow{2}{*}{mCFS}
 &RMSE &3.07&4.08&3.20\\
 &2SD &6.07&8.15 &6.36\\
 \hline
 \multirow{2}{*}{UDFS}
 &RMSE &3.59&4.58&4.08\\
 &2SD &7.18&9.01 &8.13\\
 \hline
 \multirow{2}{*}{Llcfs}
 &RMSE &2.90&4.21&3.05\\
 &2SD &6.30&8.39 &6.07\\
 \hline
 \multirow{2}{*}{CFS}
 &RMSE &3.10&4.41&3.19\\
 &2SD &6.30&8.83 &6.37\\
 \hline
 \multirow{2}{*}{Fsasl}
 &RMSE &2.80&4.04&3.09\\
 &2SD &5.60&8.07 &6.17\\
 \hline
 \multirow{2}{*}{Ufsol}
 &RMSE &2.79&4.04&3.09\\
 &2SD &5.59&8.07 &6.17\\
 \hline
 \multirow{2}{*}{Lasso}
 &RMSE &2.72&3.78&2.73\\
 &2SD &5.45&7.52 &5.45\\
 \hline
  \multirow{2}{*}{Fitrgp}
 &RMSE &\textbf{2.57}&\textbf{3.60}&\textbf{2.58}\\
 &2SD &\textbf{5.13}&\textbf{7.20} &\textbf{5.14}\\
 \hline
 \end{tabular}
 \end{adjustbox}
\label{result fsa2}
\end{table}

\begin{figure}[pos = H]
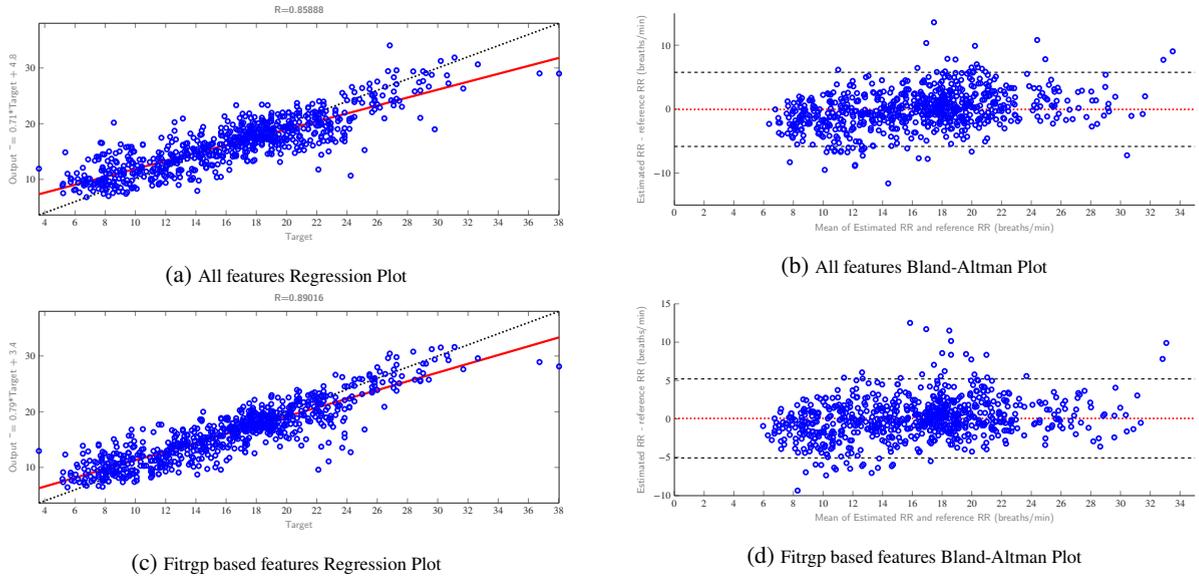

    \begin{subfigure}{0.45\textwidth}
    \begin{adjustbox}{max width=1\textwidth,center}
    \input{figs/all_feat_reg}
    \end{adjustbox}
    \caption{\scriptsize All features Regression Plot}
    \end{subfigure}
    \hspace{0.04\textwidth}
    \begin{subfigure}{0.45\textwidth}
    \begin{adjustbox}{max width=1\textwidth,center}
    \input{figs/all_feat_bland}
    \end{adjustbox}
    \caption{\scriptsize All features Bland-Altman Plot}
    \end{subfigure}

    \begin{subfigure}{0.45\textwidth}
    \begin{adjustbox}{max width=1\textwidth,center}
    \input{figs/9_reg}
    \end{adjustbox}
    \caption{\scriptsize Fitrgp based features Regression Plot}
    \end{subfigure}
    \hspace{0.04\textwidth}
    \begin{subfigure}{0.45\textwidth}
    \begin{adjustbox}{max width=1\textwidth,center}
    \input{figs/9_band}
    \end{adjustbox}
    \caption{\scriptsize Fitrgp based features Bland-Altman Plot}
    \end{subfigure}
    \caption{GPR model with  Regression and Bland-Altman plot (a-b) for all features and (c-d) for Fitrgp based features.}
    \label{fig abcd}
\end{figure}

\textbf{Figure \ref{fig abcd}}, shows the best performing GPR model with and without the use of the feature selection algorithm. The result is visualized using regression and a Bland-Altman plot. The regression plot allows seeing how close the predictions are to the ground truth with the help of a trendline. The closer the trendline is to the $y = x$line, the better the model. Bland-Altman plot allows us to see the spread of the data and also allows us to see the 95\% limit of agreement (LOA) of the data, where a smaller LOA means a better model. \textbf{Figure \ref{fig abcd}}, shows that with all features, the algorithm had an R-value of 0.859 and an LOA of -5.83 to 5.76 bpm. With the feature selection algorithm (Fitrgp), the R-value is increased to 0.890 and the LOA reduces to -5.08 to 5.21 bpm.\\
\begin{figure}[pos = H]
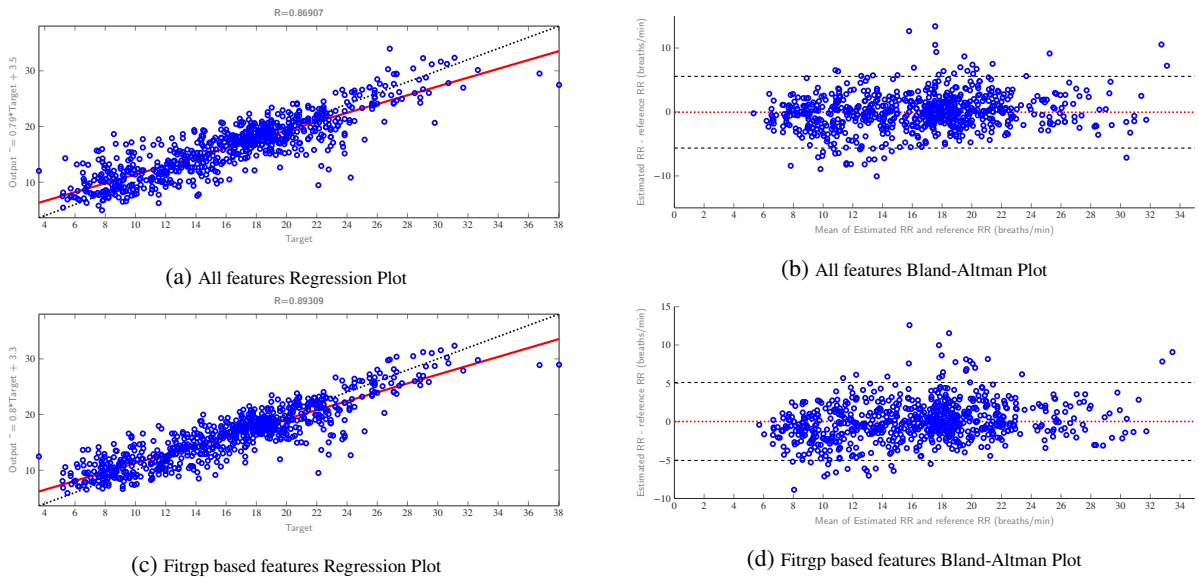

    \begin{subfigure}{0.45\textwidth}
    \begin{adjustbox}{max width=1\textwidth,center}
    \input{figs/all_feat_opt_reg}
    \end{adjustbox}
    \caption{\scriptsize All features Regression Plot}
    \end{subfigure}
    \hspace{0.04\textwidth}
    \begin{subfigure}{0.45\textwidth}
    \begin{adjustbox}{max width=1\textwidth,center}
    \input{figs/all_feat_opt_bland}
    \end{adjustbox}
    \caption{\scriptsize All features Bland-Altman Plot}
    \end{subfigure}

    \begin{subfigure}{0.45\textwidth}
    \begin{adjustbox}{max width=1\textwidth,center}
    \input{figs/9_opt_reg}
    \end{adjustbox}
    \caption{\scriptsize Fitrgp based features Regression Plot}
    \end{subfigure}
    \hspace{0.04\textwidth}
    \begin{subfigure}{0.45\textwidth}
    \begin{adjustbox}{max width=1\textwidth,center}
    \input{figs/9_opt_band}
    \end{adjustbox}
    \caption{\scriptsize Fitrgp based features Bland-Altman Plot}    
    \end{subfigure}
    \caption{Optimized GPR model with  Regression and Bland-Altman plot (a-b) for all features and (c-d) for Fitrgp based features.}
    \label{fig efgh}
\end{figure}
The effect of hyperparameter optimization of the GPR model is shown in \textbf{Figure \ref{fig efgh}}. When comparing both the optimized models, it can be seen that the model with no feature selection had an R-value of 0.869 and an LOA of -5.65 to 5.56 bpm while the optimized GPR model with feature selection provides an R-value of 0.893 and the LOA of -5.05 to 5.12 bpm. Hence, it can be concluded that the feature selection algorithm helps in increasing the performance of the GPR model. Comparing \textbf{Figures \ref{fig abcd}} and \textbf{\ref{fig efgh}}, it can be noticed that hyperparameter tuning has helped both the models. However, the best performance can be observed with the optimized GPR along with the Fitrgp feature selection algorithm.
\begin{table}[pos = H]
\centering
\caption{Comparison of the proposed method with the recent related works concerning the database, methodology, and estimation error}
\begin{adjustbox}{max width=1\textwidth,center}
 \begin{tabular}{||p{0.11\textwidth}| p{0.06\textwidth} | p{0.13\textwidth} |p{0.10\textwidth} |p{0.13\textwidth} |p{0.10\textwidth} |p{0.15\textwidth} ||}
\hline
\textbf{Author} &\textbf{Year} &\textbf{Database}&\textbf{Subject} & \textbf{Method} &\textbf{Metric} &\textbf{Result}\\ [0.5ex] 
 \hline\hline
\multirow{5}{0.1\textwidth}{Motin et al.\cite{motin2020}} &&&&& MAE &3.05 \\ && Own & 10 & Empirical & RMSE &- \\  & 2020 &  Database & Subjects & Mode & R &- \\ && &&  Decom- & 2SD &-\\ && && position & LOA &- \\
 \hline
 \multirow{5}{0.1\textwidth}{L'Her et al.\cite{l2019}} &&&&& MAE &- \\ && Own & 30 && RMSE &- \\  & 2019 &  Database & ICU & Own & R &0.78 \\ && & Patient & Approach & 2SD &- \\ &&&&& LOA &- \\ 
\hline
 \multirow{5}{0.1\textwidth}{Motin et al.\cite{motin2019}} &&&&& MAE &0 - 5.03 \\ && MIMIC & 53 & Empirical & RMSE &- \\  & 2019 &  Database & Subjects & Mode & R &- \\ && &&  Decom- & 2SD &- \\ && && position & LOA &- \\
 \hline
 \multirow{5}{0.1\textwidth}{Jarch et al.\cite{jarchi2018}} &&&&& MAE &2.56 \\ && BIDMC & 10 & Accelero- & RMSE &- \\  & 2018 &  Dataset & Subjects & meter & R &- \\ && & && 2SD &- \\ &&&&& LOA &- \\
 \hline
 \multirow{5}{0.1\textwidth}{Pirhonen et al.\cite{pirhonen2018}} &&&&& MAE &2.33 \\ && Vortal & 39 & Wavelet & RMSE &3.68 \\  & 2018 &  Database & Subjects & Synchro- & R &- \\  && & & squeezing & 2SD &- \\ && & & Transform & LOA &- \\
 \hline
 \multirow{5}{0.1\textwidth}{Zhang et al.\cite{zhang2017}} && & & Joint & MAE &- \\ && Capnobase & 42 & sparse & RMSE &2.89 \\  & 2017 &  Dataset & Subjects & signal & R &- \\ && & & Reconstr- & 2SD &5.23 \\ && & & action & LOA &-5.6 to 4.9 \\
 \hline
 \multirow{5}{0.1\textwidth}{Charlton et al.\cite{charlton2016}} &&&&& MAE &- \\ && Vortal & 39 & 92 & RMSE &- \\  & 2016 &  Dataset & Subjects & Different & R &- \\ && & & Algorithm & 2SD &6.20 \\ &&&&& LOA &-5.2 to 7.2 \\
 \hline
 \multirow{5}{0.1\textwidth}{This Work} &&&& & MAE &\textbf{1.91} \\ && Vortal & 39 & Machine & RMSE &\textbf{2.57} \\  & 2020 &  Database & Subjects & Learning & R &\textbf{0.89} \\  && & && 2SD &\textbf{5.13} \\&&&&& LOA &\textbf{-5.0 to 5.1} \\[1ex]
\hline
\end{tabular}
\end{adjustbox}
\label{result}
\end{table}

Several factors made it difficult to compare the reported performance of algorithms of different research groups in the literature, such as the use of different statistical tests, data from different subject groups, and the lack of consistent algorithm implementations. As a result, it is not possible to decide from the literature which algorithms score higher. A comprehensive comparison of RR estimation is summarized with the state-of-the-art literatures in \textbf{Table \ref{result}}. As shown in \textbf{Table \ref{result}} Motin et al.\cite{motin2020} introduced a novel approach to the continuous control of PPG-based RR estimation using a smart fusion strategy based on EEMD is one of the best performing approaches. Estimating RR under daily living conditions is challenging, as the PPG signal is affected by the motion artifacts. The median absolute error (MAE) in \cite{motin2020}. L’Her et al.\cite{l2019} described the accuracy of measurements of the respiratory rate using a specially developed reflex-mode photoplethysmographic pathological signal analysis (PPG-RR) and validated its implementation within medical devices. They experimented with this on 30 intensive care unit (ICU) patients where a correlation coefficient for RR of 0.78 was achieved. Motin et al. \cite{motin2019} used the EMD family and PCA-based hybrid model to remove RR from PPG, a natural extension of their previously 

\noindent built hybrid PCA-EMD (EEMD) system. The MAE for the model tested on MIMIC datasets of 53 subjects were varied from 0 to 5.03 bpm.\\

With the assistance of Time Frequency (TF) reassignments and a particle filter, Pirhonen et al.\cite{pirhonen2018} suggested the use of amplitude fluctuations of the PPG signals to approximate RR. Vortal database was used in that study. The highest results were achieved using wavelet synchrosqueezing transform, which produced an MAE and RMSE of 2.33 and 3.68 bpm, respectively. Jarchi et al.\cite{jarchi2018} presented a case study on 10 subjects to estimate RR from the PPG signals relative to accelerometry and achieved an MAE of 2.56 bpm. Zhang et al.\cite{zhang2017} proposed the estimation of the RR from the PPG signal using joint sparse signal reconstruction (JSSR) and Spectra Fusion (SF) from 42 subjects and achieved an LOA, 2SD, and RMSE of -5.58 to 4.88, 5.23, 2.81 bpm, respectively. Charlton et al.\cite{charlton2016} divided the algorithm into three phases: respiratory signal extraction, RR estimation, and estimation fusion and 314 different algorithms were assessed and the best algorithm had 95 percent LOA and 2SD of -5.1 to 7.2, and 6.2 bpm, respectively.\\

There is no exact medical standard regarding the estimation of RR. However, in a review paper\cite{charltonreview2017} where over 196 traditional RR extraction technique were reviewed, they stated that an MAE less than 2 bpm should provide a suitable indicator for a good estimator. The machine learning model suggested in this analysis was measured with much higher precision and accuracy which shown in \textbf{Table \ref{result}}

\section{Conclusions}
\label{Conclusions}
\hspace{\parindent}In this study, the authors proposed and developed a machine learning-based method for predicting RR from the PPG signal features. This successfully shows how the PPG signal can be used to correctly estimate the RR value invasively. The entire prepossessing process of the PPG signals to extract the features, feature selection, and training of the algorithms were discussed. The method used 107 time-domain, frequency-domain, and statistical features to extract meaningful information from the PPG signal. ANN and GRNN and 19 other machine learning models were trained, validated, and tested for RR estimation, where the performance of GPR, SVR, ensemble trees, ANN and GRNN were promising. To reduce computational complexity and the risk of over-fitting, different feature selection algorithms were investigated. It was observed that a combination of Fitrgp feature selection and GPR machine-learning algorithm produced the best result. However, hyper-parameter optimization can improve the model performance further. The fine-tuned model provides an RMSE, MAE, R, and 2SD score of 2.57, 1.91, 0.89, and 5.13 bpm for the estimation of RR. This state-of-the-art performance of the proposed model will make it possible to deploy this for ambulatory and intensive care units as well for remote health care monitoring.
\section*{Author Contributions}
 Experiments were designed by MNIS, MHC and MEHC. Experiments were performed by MNIS, MHC and MEHC. Results analysis, and interpretation and paper drafting were done by all authors. 
\section*{Acknowledgement}
\label{Acknowledgement}
This work was made possible by NPRP12S-0227-190164 from the Qatar National Research Fund, a member of Qatar Foundation, Doha, Qatar. The statements made herein are solely the responsibility of the authors.

\bibliographystyle{elsarticle-num}

\bibliography{ppg}

\end{document}